\documentclass[journal]{IEEEtran}

\usepackage{cite}
\usepackage{amsmath,amssymb,amsfonts}
\usepackage{algorithmic}
\usepackage{graphicx}
\usepackage{textcomp}

\usepackage{multirow}
\usepackage{array}
\usepackage{caption}
\usepackage{subcaption}
\usepackage{enumerate}
\usepackage{booktabs}
\usepackage{threeparttable}
\usepackage{verbatim}
\usepackage{xcolor}
\usepackage{hyperref}
\usepackage{makecell}

\def\BibTeX{{\rm B\kern-.05em{\sc i\kern-.025em b}\kern-.08em
		T\kern-.1667em\lower.7ex\hbox{E}\kern-.125emX}}

\begin{document}
	\title{The Lighter The Better: Rethinking Transformers in Medical Image Segmentation Through Adaptive Pruning}
	\author{Xian Lin, Li Yu, Kwang-Ting Cheng, and Zengqiang Yan
		\thanks{\textit{Corresponding author: Zengqiang Yan}}
		\thanks{X. Lin, L. Yu, and Z. Yan are with the School of Electronic Information and Communications, Huazhong University of Science and Technology, Wuhan 430074, China (e-mail: z\_yan@hust.edu.cn).}
		\thanks{K. -T. Cheng is with the School of Engineering, Hong Kong University of Science and Technology, Kowloon, Hong Kong.}
	}
	
	\maketitle
	
	\begin{abstract}
		
		Vision transformers have recently set off a new wave in the field of medical image analysis due to their remarkable performance on various computer vision tasks. However, recent hybrid-/transformer-based approaches mainly focus on the benefits of transformers in capturing long-range dependency while ignoring the issues of their daunting computational complexity, high training costs, and redundant dependency. In this paper, we propose to employ adaptive pruning to transformers for medical image segmentation and propose a lightweight and effective hybrid network APFormer. To our best knowledge, this is the first work on transformer pruning for medical image analysis tasks. The key features of APFormer mainly are self-supervised self-attention (SSA) to improve the convergence of dependency establishment, Gaussian-prior relative position embedding (GRPE) to foster the learning of position information, and adaptive pruning to eliminate redundant computations and perception information. Specifically, SSA and GRPE consider the well-converged dependency distribution and the Gaussian heatmap distribution separately as the prior knowledge of self-attention and position embedding to ease the training of transformers and lay a solid foundation for the following pruning operation. Then, adaptive transformer pruning, both query-wise and dependency-wise, is performed by adjusting the gate control parameters for both complexity reduction and performance improvement. Extensive experiments on two widely-used datasets demonstrate the prominent segmentation performance of APFormer against the state-of-the-art methods with much fewer parameters and lower GFLOPs. More importantly, we prove, through ablation studies, that adaptive pruning can work as a plug-n-play module for performance improvement on other hybrid-/transformer-based methods. Code is available at \url{https://github.com/xianlin7/APFormer}.	
		
	\end{abstract}
	
	
	\section{Introduction}
	\label{sec:introduction}
	
	\IEEEPARstart{T}{ransformer}, a new type of neural networks, originally designed for natural language processing (NLP) \cite{d1}, has achieved remarkable performance on various computer vision tasks \cite{d2, d3, d4, d5}, breaking the longstanding dominance of convolutional neural networks (CNNs) \cite{d6, d7, d8, d40}. 
	
	Compared to CNNs benefiting from translation equivariance, sparse interaction, and weight sharing while suffering from limited receptive fields, transformers reveal extraordinary ability in capturing long-range dependency with a multi-head self-attention module, which is more consistent with the human vision system in feature extraction \cite{d43, d44, d45, d46, d47}. Therefore, transformers have drawn explosive attention in medical image segmentation. Chen \textit{et al.}\cite{d9} proposed TransUNet by introducing transformers into the encoder of U-Net, which is the first exploration of transformers in medical image segmentation. Zhang \textit{et al.}\cite{d10} developed a novel parallel-in-branch architecture to combine CNNs and transformers to make full use of their respective advantages. Karimi \textit{et al.}\cite{d11} proposed the first convolution-free model relying only on self-attention and multilayer perceptron for medical image segmentation and achieved competitive results. Wang \textit{et al.}\cite{d12} designed TransBTS to make convolution and transformer responsible for local and global feature capturing respectively and complement each other. Chen \textit{et al.}\cite{d13} proposed TransAttUnet, a CNN-Transformer hybrid model with multi-level attention, to effectively address the information recession problem caused by the limited interaction range of CNNs. Though these transformer-based/hybrid methods have achieved encouraging results for medical image segmentation, they still suffer from poor convergence on small-scale datasets and daunting computational complexity, which in turn hinders the deployment of transformers in clinical scenarios.
	
	To alleviate severe performance degradation of transformers when deployed on small-scale datasets, several methods have been proposed to build a competitive transformer-based model without training/pre-training on large-scale datasets. Xie \textit{et al.}\cite{d14} developed CoTr to speed up the convergence with a deformable self-attention mechanism to focus on partial areas. Gao \textit{et al.}\cite{d15} constructed a U-shaped CNN-Transformer hybrid model to avoid pre-training by making a mass of convolutional operations responsible for capturing local features. Valanarasu \textit{et al.}\cite{d16} devoted to deploying transformers on small-scale datasets by introducing learnable gate parameters to restrain the effect of poor position embedding. Wang \textit{et al.}\cite{d17} proposed boundary-aware transformers for performance improvement by merging boundary-wise prior knowledge. Gao \textit{et al.}\cite{d18} applied depth-wise separable convolution on the linear projection of self-attention and the feed-forward parts of the transformer to allow the model to be trained from scratch on small-scale datasets. Though these methods ease the deployment of transformers on small-scale medical image datasets by introducing inductive biases, the high computational complexity issue still has not been properly addressed.
	
	Approaches for computation reduction of medical transformers can be classified into three categories: dependency establishment based on windows, deformable self-attention, and scale downsampling of queries and keys. Window-based models are mainly inspired by Swin Transformer\cite{d3}, whose representations are extracted within shifted windows. Cao \textit{et al.}\cite{d11} built an UNet-like architecture only with Swin Transformer blocks to form a pure transformer-based model with acceptable computational cost. Similarly, Li \textit{et al.}\cite{d19} presented a novel upsampling method by introducing Swin Transformer blocks to the decoder of U-Net. Though encouraging performance has been achieved by these methods, their local features are seriously destroyed by the rigid window partitioning scheme in Swin Transformer. To address this, Lin \textit{et al.}\cite{d20} adopted a double-scale encoder to restore local information loss between different-scaled Swin Transformers. Deformable self-attention methods directly reduce computational complexity by only building dependency within a subset of patches \cite{d14, d21}. These two types of approaches ease the computation burden of transformers at the risk of losing important long-range dependency, which may result in sub-optimal feature expression. Comparatively, downsizing the scale of queries and keys by pooling\cite{d15}, grouping\cite{d22}, and dilating\cite{d23} can sufficiently preserve long-range dependency since they are uniformly downscaled at equal intervals. However, uniform scaling ignores the difference in importance of query-key pairs and cannot build optimal dependence relationships.
	
	Transformer pruning, a powerful approach to building lightweight transformers by dropping redundant structures or components, can effectively address the above issues. To date, there has not been any study focusing on exploring the potential benefit of transformer pruning for medical image segmentation. In fact, transformer pruning is under-explored even in the natural image domain. Zhu \textit{et al.}\cite{d24} presented the first vision transformer pruning method by evaluating the impact of transformer dimensions and dropping unimportant dimensions accordingly. Pan \textit{et al.}\cite{d25} developed an interpretable module to adaptively filter out the redundant patches. Rao \textit{et al.}\cite{d26} introduced a prediction module to score each patch and then pruned redundant patches hierarchically. Yin \textit{et al.}\cite{d27} reduced the inference cost by automatically minimizing the number of tokens. Despite the great results achieved by these approaches, they only focused on the classification/recognition tasks and reduced the computational complexity at the cost of minor performance degradation.
	
	In this paper, we, for the first time, propose a transformer pruning framework APFormer for lighter and better medical image segmentation. Specifically, a U-shaped CNN-Transformer hybrid framework is first constructed as the base framework for capturing sufficient local and global information. For the transformer module, three components are developed to make it feasible to train efficient transformers from scratch with fewer rounds: self-supervised self-attention (SSA), Gaussian-prior relative position embedding (GRPE), and adaptive pruning. SSA introduces the well-converged dependency distribution as prior knowledge to establish high-quality dependency relationships. GRPE highlights the nearby patches for each query patch when introducing position information. These two components help transformers converge well without relying on expensive training data and time. Based on this transformer framework, transformer pruning follows a two-stage process. In the first stage, pruning is accomplished by dropping those query patches with high confidence of being background, which can significantly reduce redundant computation in useless queries. As for the second stage, redundant dependencies are dynamically eliminated for each query, which can decrease the perceptual redundancy of each crucial patch. Qualitative and quantitative results compared to the state-of-the-art approaches on two public datasets demonstrate the effectiveness of APFormer for medical image segmentation. More importantly, the proposed transformer pruning method can be extended to other hybrid-/transformer-based methods for computational complexity reduction and performance improvement. The contributions can be summarized as follows:
	
	\begin{itemize}
		
		\item APFormer, a lightweight CNN-Transformer hybrid segmentation framework for medical image segmentation, which is more feasible for deployment.
		
		\item A two-stage transformer pruning mechanism for simultaneous computational complexity reduction and performance improvement. To our best knowledge, this is the first work on transformer pruning for image segmentation. Furthermore, in contrast to existing transformer pruning methods which inevitably suffer from performance degradation, the proposed pruning mechanism can achieve performance improvement.
		
		\item A self-supervised strategy for improving self-attention matrices in transformers to build truly-useful long-range dependency faster.
		
		\item A novel position embedding approach utilizing prior knowledge to better introduce position information with lower training complexity.
		
	\end{itemize}
	
	\begin{figure*}[!t]
		\centerline{\includegraphics[width=1\textwidth]{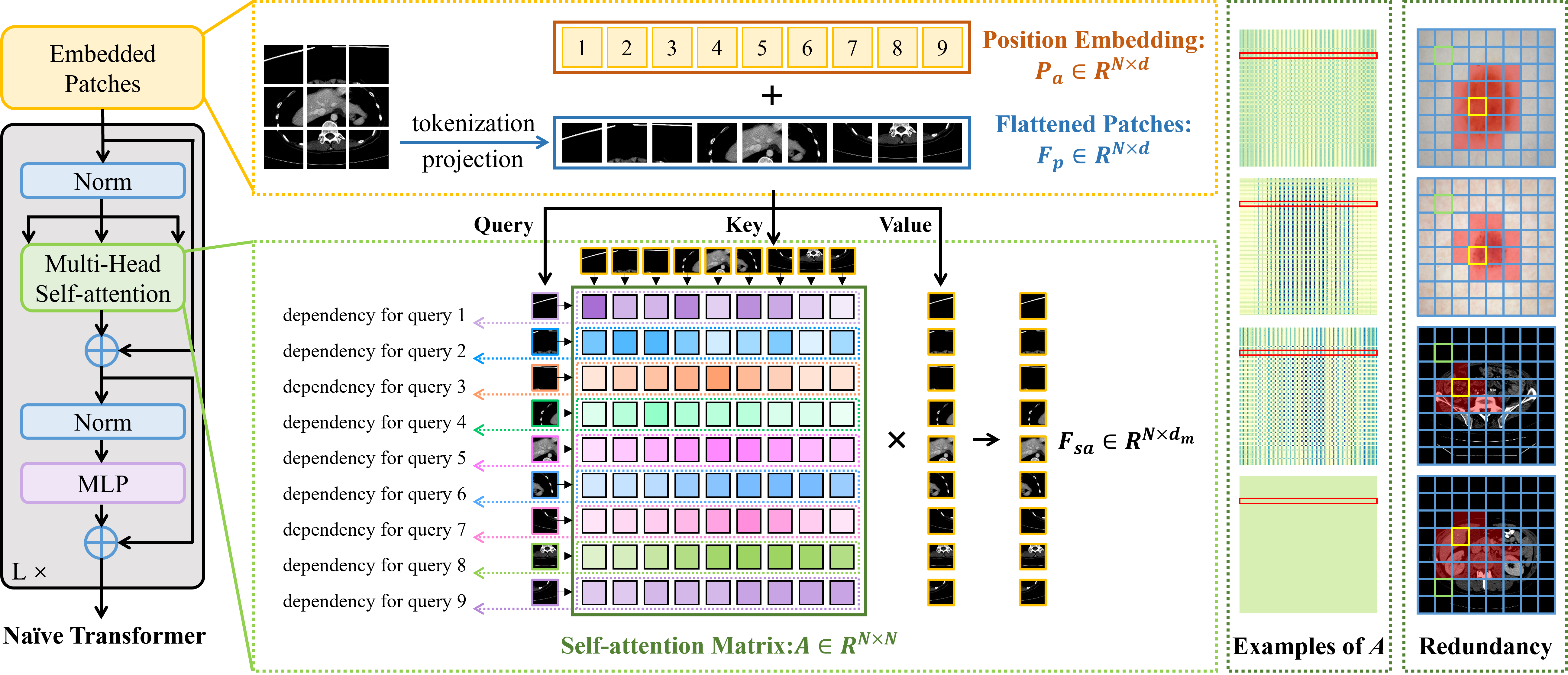}}
		\caption{
			Analysis of challenges in deploying transformers for medical image segmentation.
		}
		\label{fig1}
	\end{figure*}

	\section{Problem Analysis}
	
	In this section, we revisit the structure of naive vision transformers and analyze the challenges in deploying them for medical image segmentation.
	
	\subsection{Structure of Naive Vision Transformer}
	
	Typical vision transformer is composed of a multi-head self-attention (MSA) block and a feed-forward network (FFN) in series as shown in Fig.~\ref{fig1}. Given the input feature $F \in \mathbb{R} ^ {D \times H \times W}$, it is tokenized and projected into a sequence of flattened patches $F_p \in \mathbb{R} ^ {N \times d}$, where $N$ is the number of patches. Since the position information of $F$ is corrupted during tokenization, an input-independent learnable absolute position embedding $P_{a} \in \mathbb{R} ^ {N \times d}$ is utilized to compensate for the lost position information in $F_p$. Specifically, $F_p+P_{a}$ is taken as the input of MSA and projected into query $Q \in \mathbb{R} ^ {N \times d_m}$, key $K \in \mathbb{R} ^ {N \times d_m}$, and value $V \in \mathbb{R} ^ {N \times d_m}$ to build global dependency, which can be summarized as
	\begin{equation}
	\begin{array}{*{20}{c}}
	{F_{sa} = A(F_p+P_{a})E_v}\\
	{F_{msa} = (F_{sa_1} \textcircled{c} \cdot \cdot \cdot \textcircled{c} F_{sa_h})W_{msa} \textcircled{+} (F_p + P_{a})},
	\end{array}
	\label{eq1}
	\end{equation}
	where $A= softmax(\frac{(F_p+P_{a})E_q((F_p+P_{a})E_k)^T}{\sqrt{d_m}}) \in \mathbb{R} ^ {N \times N}$ is the attention matrix, $h$ is the number of self-attention heads, $\textcircled{c}$ and $\textcircled{+}$ are the concat and residual connection operations, $E_q \in \mathbb{R} ^ {d \times d_m}$, $E_k \in \mathbb{R} ^ {d \times d_m}$, and $E_v \in \mathbb{R} ^ {d \times d_m}$ are the projection matrices for generating $Q$, $K$, and $V$ from the self-attention input, and $W_{msa}$ is the projection matrix for combining the output of multiple heads. Then, the final learned features of the transformer module can be obtained by applying FFN to $F_{msa}$ by
	\begin{equation}
	F_{ffn} = (max(0, F_{msa}W_{1}+b_{1}) \cdot W_{2}+b_{2}) \textcircled{+} F_{msa},
	\label{eq2}
	\end{equation}
	where $W_{1}$ and  $W_{2}$ are learnbale projection matrices and $b_{1}$, $b_{2}$ are the offsets in FFN. Obviously, the main benefit of vision transformers comes from the construction of $A$ which builds patch-wise interactions for all to capture long-range, global dependency.

	\begin{figure*}[!t]
		\centerline{\includegraphics[width=1\textwidth]{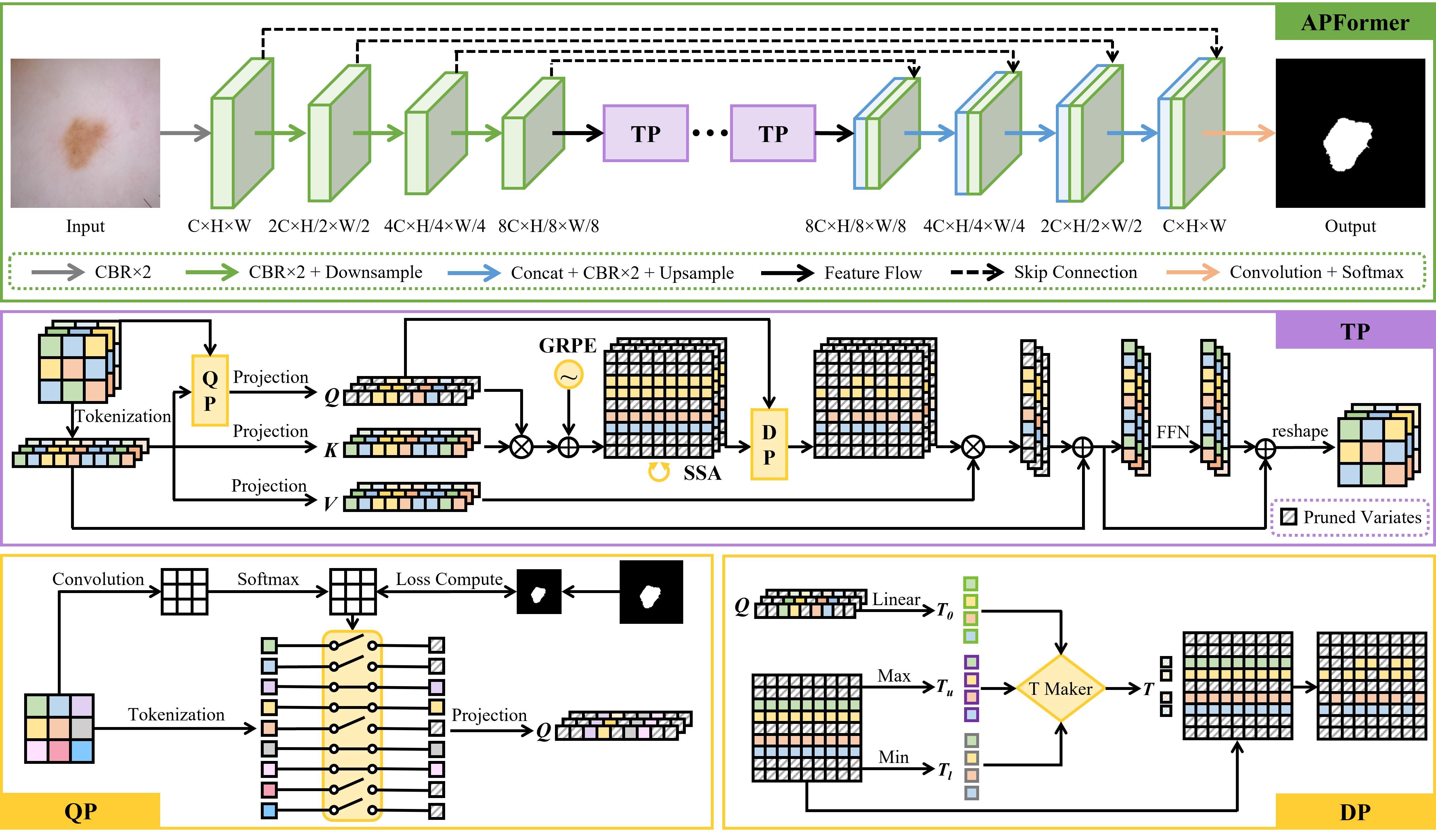}}
		\caption{Overview of APFormer for medical image segmentation based on transformer pruning, where TP, SSA, GRPE, QP, and DP represent transformer pruning, self-supervised self-attention, Gaussian-prior relative position embedding, query pruning, and dependency pruning respectively.}
		\label{fig2}
	\end{figure*}

	\subsection{Challenges of Transformer Deployment for Medical Image Data}
	The success of transformers in the natural image domain relies on sufficient training time and data. However, when applying transformers to medical image segmentation, severe performance degradation is encountered due to:
	\begin{itemize}
		
		\item \textbf{Expensive training costs in learning self-attention matrices and position embedding:} As presented in Fig.~\ref{fig1}, each self-attention matrix $A$ can be regarded as $N$ convolution kernels with the same size $h \times w$ to capture features for each query patch based on global information, where $h \times w$ is the number of patches (\textit{i.e.} each value in $(h,w)$ represents a patch). Here, $N=h \times w=H/s \times W/s$ is determined according to the size of each patch $(s,s)$ (\textit{e.g.} $s=2^0,2^1,2^2,2^3,2^4$ in practice). Consequently, the calculation of $A$ (\textit{i.e.}, $N \times N$) is highly expensive (\textit{e.g.} $N=256,1024$ in practice), requiring a large amount of data for training. In addition, position embedding, which is introduced for compensating the spatial location information destroyed by tokenizing 2D features into 1D sequences, is a $N \times d$ matrix, the scale and irregularity of which further exacerbates the data dependence problem. Unfortunately, different from natural images, collecting large-scale well-annotated medical images is infeasible due to data privacy and high annotation cost, making it challenging to guarantee good convergence of both self-attention and position embedding. Based on the examples of $A$ sampled from four layers of transformers trained on the ISIC 2018 dataset \cite{d41,d42} from scratch, illustrated in the second last column of Fig.~\ref{fig1}, the dependency built for each query (\textit{e.g.}, each highlighted row of $A$) is quite similar within each of the four self-attention matrices, indicating the poor performance of $A$ on building truly-important/-specific dependency for each query patch.
		
		\item \textbf{Severe computation and dependency redundancy in self-attention:} Transformers are known for their outstanding ability to build long-range dependency. However, as presented in Fig.~\ref{fig1}, each patch would be taken as a query (\textit{i.e.}, one row of $A$) and then build dependency with all patches (\textit{i.e.}, all columns of $A$). Though building long-range dependency is beneficial, involving all patches could be highly redundant and would bring non-negligible noise due to perception redundancy. For example, as shown in the last column of Fig.~\ref{fig1}, building long-range dependency for the green patches (\textit{i.e.} easy-to-identify background patches) is worthless, resulting in high computational redundancy. Meanwhile, for the patches in yellow, interactions with the patches highlighted in red via self-attention are beneficial, while the information from other patches can be redundant and even noisy.
		
	\end{itemize}
	
	Based on the above analysis, with the assistance of well-converged self-attention matrices and position embedding, pruning redundant dependency in transformers can be beneficial even for performance improvement in addition to computation cost reduction.
	
	\section{Method}
	
	APFormer is constructed by first building a pruning-friendly CNN-Transformer hybrid architecture and then performing adaptive pruning. Details are provided in the following.
	
	\subsection{Overview}
	
	APFormer is constructed as a simple U-shaped structure with pruned transformer blocks as the bridge between encoder and decoder as depicted in Fig.~\ref{fig2}. Given an 2D input medical image $X \in \mathbb{R} ^ {1 \times H \times W}$ or $X \in \mathbb{R} ^ {3 \times H \times W}$ with the resolution of $H \times W$, the CNNs encoder extracts shallow and fine features to capture short-range dependency and builds inner inductive biases. Then, the pruned transformer modules model long-term interactions with lower computation and perception redundancy. Finally, the decoder produces the final segmentation results.
	
	\subsection{Construction of SSA and GRPE}
	
	The key to lossless pruning is to effectively identify and remove redundancy while preserving all useful information. Consequently, before pruning, we propose self-supervised self-attention (SSA) and Gaussian-prior relative positional embedding (GRPE) to achieve better convergence and to better recognize redundancy for lower training cost.
	
	\subsubsection{Self-Supervised Self-Attention}
	
\begin{figure}[!t]
		\centerline{\includegraphics[width=1\columnwidth]{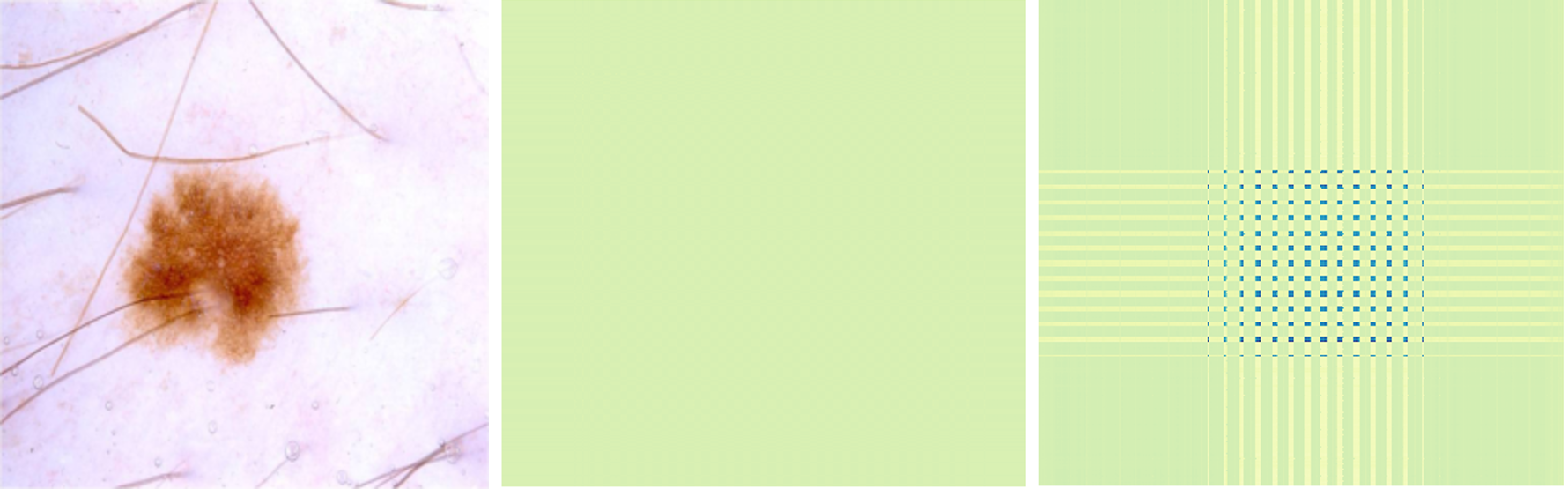}}
		\caption{Examples of self-attention matrices. Left: the input medical image. Middle: a badly-converged self-attention matrix that treats all patches equally. Right: a well-converged self-attention matrix that is consistent with human's intuitive attention.}
		\label{fig3}
	\end{figure}

	The design of SSA is inspired by our observation and analysis of well-converged self-attention matrices. Specifically, two extreme examples of self-attention matrices are presented in Fig.~\ref{fig3}, where the left one generally happens in the deep layers of transformers due to feature collapse and the right one is a well-converged attention matrix. It can be seen that, in medical image segmentation, a well-converged attention matrix should have the following characteristics: 
	\begin{itemize}
		\item Strong dependencies are compact, mainly around the diagonal regions with the foreground patches. It is consistent with the fact that dependency is inversely proportional to the distance between patches and only built among interesting regions.
		\item Attention matrices are roughly symmetric (not completely symmetric). It is consistent with the fact that the dependency between each pair of patches usually is mutual.
		\item Attention matrices are of low entropy. It is consistent with clinical prior knowledge that long-range dependency among structures/organs/tumors should be relatively stable with minimal uncertainty.
	\end{itemize}
	Accordingly, for better training, given an attention matrix $A$, two constraints are constructed to supervise the learning of the transformer module. One is on the diagonal symmetry of $A$, defined as
	\begin{equation}
	\mathcal{S}(A, A^T) = \frac{\sum_{i=0}^{N}\sum_{j=0}^{N}(A_{i, j} \times A_{i, j}^T)}{\left\|A\right\|_2 \times \left\|A^T\right\|_2},
	\label{eq3}
	\end{equation}
	where $A^T$ is the transpose of $A$ and $\left\| \cdot \right\|_2$ is the operation of calculating quadratic normal form. Then, the symmetry loss $L_{sym}$ is defined as
	\begin{equation}
	L_{sym} = Max(1 - \mathcal{S}(A, A^T) - \alpha_{sym}, 0),
	\label{eq4}
	\end{equation}
	where $\alpha_{sym}$ is a smoothing factor, as pursuing $A$ and $A^T$ being identical can be counter-productive.
	The other constraint is on the entropy of $A$, where its entropy is obtained by
	\begin{equation}
	\mathcal{E}(A_i) = \frac{-\sum_{j=0}^{N}(A_{i, j} \times \log_2(A_{i, j}))}{\log_2(N)}.
	\label{eq5}
	\end{equation}
	To pursue low entropy, a entropy loss $L_{en}$ is defined as
	\begin{equation}
	L_{en} = Max(Min(\mathcal{E}(A_i)) - \alpha_{en}, 0),
	\label{eq6}
	\end{equation}
	where $\alpha_{en}$ is a smoothing factor to release the penalty. Without $\alpha_{en}$, the attention matrix would just focus on very few patches. Finally, each self-supervised self-attention head in APFormer is trained by
	\begin{equation}
	L_{ssa} = \beta_1L_{sym} + \beta_2L_{en},
	\label{eq7}
	\end{equation}
	where $\beta_1$ and $\beta_2$ are balancing hyper-parameters set as 0.8 and 0.2 respectively in our experiments. Through $L_{ssa}$, APFormer is encouraged to converge to better dependency establishment.
	
	\subsubsection{Gaussian-Prior Relative Position Embedding}
	
\begin{figure}[!t]
		\centerline{\includegraphics[width=1\columnwidth]{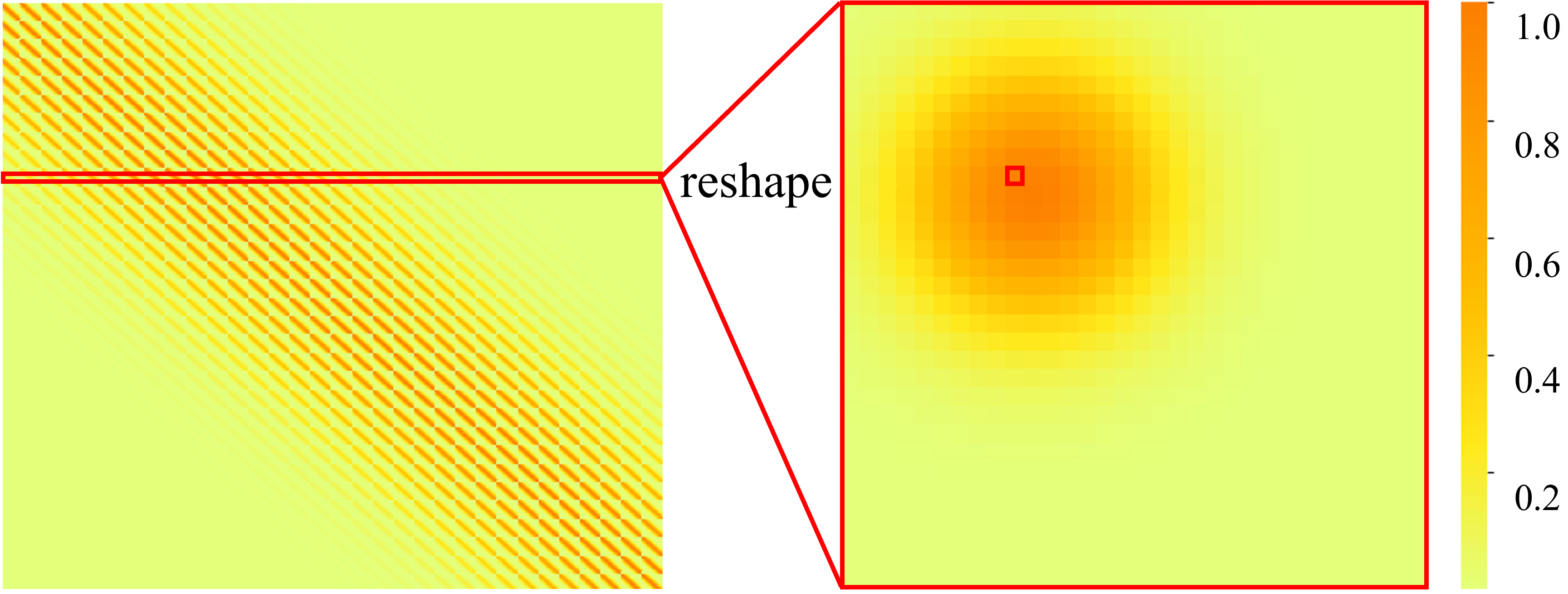}}
		\caption{Example of Gaussian relative position embedding. Left: the Gaussian relative positional embedding made from the heatmaps of all positions. Right: the Gaussian heatmap at position (10, 10) of a $32 \times 32$ feature map.}
		\label{fig4}
	\end{figure}

	Inspired by the keypoint heatmaps in pose estimation \cite{d28}, the prior importance of each key patch to the query patch can be expressed as a Gaussian distribution. As shown in Fig.~\ref{fig4}, given a $32\times32$ feature map, position $(10, 10)$ as the query patch and all positions as the key patches, the closer the key patch is to the query patch, the more important it is to the query patch. Taking each position of the feature map as a query patch orderly, the complete Gaussian heatmap with the same resolution of the self-attention matrix is as shown in Fig.~\ref{fig4}, formulated as
	\begin{equation}
	G_{i,j} = e^{-\frac{(j \textcircled{$\div$} w-i \textcircled{$\div$} w)^2 + (j \% w - i\%w)^2}{2\theta^2}},
	\label{eq8}
	\end{equation}
	where $(h, w)$ is the resolution of the feature map, $\theta$ is a learnable parameter to control the range of important regions, $\textcircled{$\div$}$ is the modulus operation, and $\%$ is the remainder operation. Then, the generated $G \in \mathbb{R} ^ {N \times N}$ can be regarded as the Gaussian prior knowledge of the relative position information. Given that it only reflects the distance relationship of relative positions while ignoring the direction information, an objective learnable relative position embedding $R \in \mathbb{R} ^ {4N}$ is introduced. The final GRPE can be obtained by
	\begin{equation}
	P_{ij} = G_{i, j} + R_{ij},
	\label{eq9}
	\end{equation}
	where $R_{ij} = R(2w(i\textcircled{$\div$}w - j\textcircled{$\div$}w + h) + (i\%w - j\%w + w))$. Compared to the number of position embedding parameters in naive vistion transformer (\textit{i.e.}, $N \times d$, generally $d=64/128/256/1024$), GPRE (\textit{i.e.}, $N \times 4 + h$, where $h$ is the number of self-attention heads and generally equals to 6/12) is much lighter.
	With the re-designed position embedding, the output of self-attention in Eq.~\ref{eq1} is rewritten as
	\begin{equation}
	F_{sa} = softmax(\frac{F_pE_qF_pE_k}{\sqrt{d_m}} + P)F_pE_v.
	\label{eq10}
	\end{equation}
	
	\subsection{Adaptive Pruning}
	
	\begin{figure*}[!t]
		\centerline{\includegraphics[width=1\textwidth]{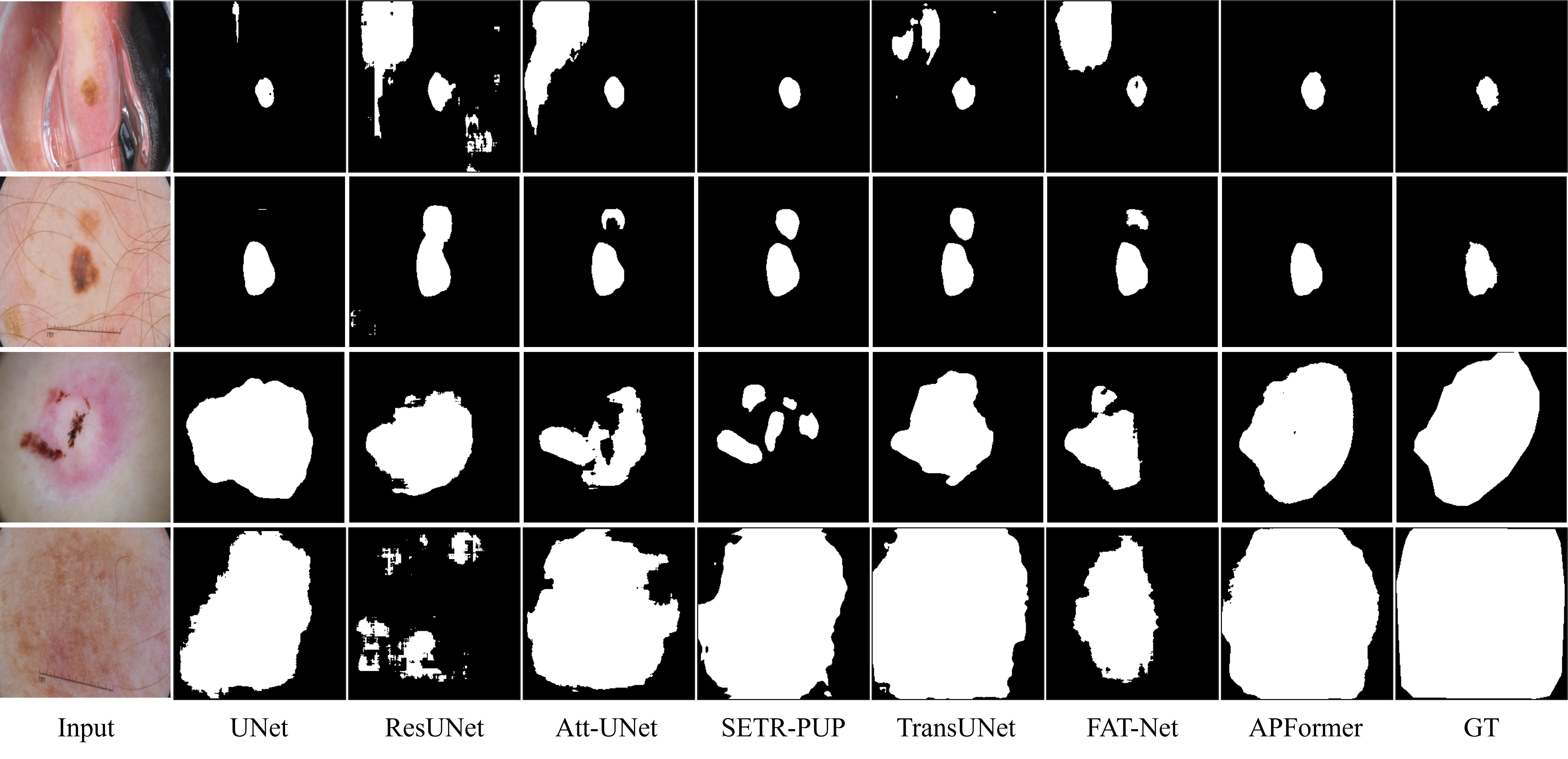}}
		\caption{Qualitative results on the ISIC dataset. From left to right: the raw images, the segmentation results produced by U-Net \cite{d6}, ResUNet \cite{d7}, Att-UNet \cite{d8}, SETR-PUP \cite{d5}, TransUnet \cite{d9}, FAT-Net \cite{d29} and APFormer respectively, and the ground truth.}
		\label{fig5}
	\end{figure*}
	{\renewcommand{\arraystretch}{1.2}
		\begin{table*}[!t]
			\centering
			\caption{Comparison results against the state-of-the-art approaches on the ISIC 2018 dataset.}\label{tab1}
			\begin{tabular}{c|c|c|cccccccc}
				\hline
				Model Type                         & Method      & Year             & Dice (\%)       & IoU (\%)        & ACC (\%)        & SE (\%)         & SP (\%)         & Params (M)    & GFLOPs       \\ \hline
				\multirow{4}{*}{CNNs}         
				& Att-UNet\cite{d8}     & 2019           & 85.66          & 77.64          & 93.76          & 86.00          & \textbf{98.26}          & 45           & 84           \\
				& CPFNet\cite{d30}       & 2020           & 87.69          & 79.88          & 94.96          & 89.53          & 96.55          & 43           & 16           \\
				& DAGAN\cite{d31}        & 2020           & 88.07          & 81.13          & 93.24          & 90.72          & 95.88          & 54           & 62           \\ 
				& CKDNet\cite{d32}       & 2021           & 87.79          & 80.41          & 94.92          & 90.55          & 97.01          & 51           & 44           \\ \hline
				\multirow{4}{*}{Transformer} & SETR-PUP$^*$\cite{d5}    & 2021    & 88.03          & 80.53          & 95.51          & \textbf{91.51}          & 96.52          & 39           & 40           \\
				& TransUnet$^*$\cite{d9}         & 2021           & 88.88          & 81.85          & 95.94          & 90.08          & 97.89         & 105         & 25         \\
				& FAT-Net\cite{d29}      & 2022           & 89.03          & 82.02          & 95.78          & 91.00          & 96.99          & 30           & 23           \\
				& \textbf{APFormer}  & 2022         & \textbf{90.70} & \textbf{84.34} & \textbf{96.50} & 90.93 & 97.81 & \textbf{2.6} & \textbf{4.1} \\ 
				\hline
			\end{tabular}
			\begin{tablenotes}
				\centering
				\item $*$ represents the re-implemented results based on released code
			\end{tablenotes}
		\end{table*}
	}
	
	To pursue lightweight and efficient transformers for medical image segmentation, we propose a two-stage transformer pruning method, including query-wise pruning and dependency-wise pruning.
	In query-wise pruning, the easy-to-identify background patches are pruned before projecting to queries as building long-range dependency for them is worthless. Specifically, we introduce a learnable gate for each candidate query patch calculated by
	\begin{equation}
	\mathcal{G}_b(q_{ij}) = \frac{e^{W_{b}F_{i,j}}}{e^{W_{b}F_{i,j}} + e^{W_{f}F_{i,j}}},
	\label{eq11}
	\end{equation}
	where $W_b$ and $W_f$ are the learnable projection matrices and are supervised by a cross-entropy loss $L_g$ between $\mathcal{G} = \mathcal{G}_b \textcircled{c} \mathcal{G}_f$ and the foreground-background ground-true mask. In query-wise pruning, only the query patches with low gate scores $F_p^p \in \mathbb{R} ^ {(1 - \alpha) N \times d}$ (\textit{i.e.}, excluding the easy-to-identify background patches) would join long-term interactions with the key patches, leading to smaller self-attention matrices $A_p \in \mathbb{R} ^ {(1 - \alpha) N \times N}$ calculated by
	\begin{equation}
	A_{p} = softmax(\frac{F_p^pE_qF_pE_k}{\sqrt{d_m}} + P).
	\label{eq12}
	\end{equation}
	
	After capturing long-range dependency for the kept query patches, we conduct dependence-wise pruning based on adaptive thresholds $T$. For each query $q_i$, its pruning threshold $T_i$ is generated from the dependency distribution and its inner features, formulated as
	\begin{equation}
	T_i = min({A_p}_i)+\frac{(max({A_p}_i)-min({A_p}_i))}{(1+e^{-W_t{F_p^pE_q}_i})(1+e^{-g})},
	\label{eq13}
	\end{equation}
	where $W_t$ is the projection matrix and $g$ is a learnable control parameter to avoid exorbitant $T_i$ at the beginning of training.  
	
	According to the adaptively-generated $T_i$, a binary decision mask $M \in \{0, 1\}^{(1 - \alpha) N \times N}$ is adopted to determine whether to prune the dependency or not. Specifically, all elements in $M$ are initialized to 1 and are updated to 0 when ${A_p}_{i, j} < T_i$ is satisfied. To make the sum of all dependency scores of each query to 1, the attention matrix $A_p$ is updated according to
	\begin{equation}
	{A_p}_{i, j} = \frac{M_{i,j}e^{{A_p}_{i, j}}}{10^{-6} + \sum_{k=0}^{N}M_{i,k}e^{{A_p}_{i, k}}}.
	\label{eq14}
	\end{equation}
	
	Through the above two-stage pruning, compared to the naive self-attention module whose computational complexity is
	\begin{equation}
	\Omega_{sa} = 3Ndd_m + 2N^2d_m + Nd_m^2,
	\label{eq15}
	\end{equation}
	the computational complexity of the pruned self-attention module in APFormer becomes
	\begin{equation}
	\Omega_{psa} = (3-\alpha)Ndd_m + \alpha(2-\lambda) N^2d_m + \alpha Nd_m^2,
	\label{eq16}
	\end{equation}
	where $\alpha \in (0, 1)$ and $\lambda \in (0, 1)$ are the query-wise pruning rate and the dependency-wise pruning rate respectively and are adaptively adjusted according to the input.

	{\renewcommand{\arraystretch}{1.2}
		\begin{table*}[!t]
			\centering
			\caption{Comparison results against the state-of-the-art approaches on the Synapse dataset. Gall, L-Kid, R-Kid, and P are the abbreviations of gallbladder, left kidney, right kidney, and parameters.}\label{tab2}
			\begin{tabular}{c|c|ccccccccccc}
				\hline
				Model Type                      & Method          & Avg.        & Aotra          & Gall   & L-Kid   & R-Kid  & Liver          & Pancreas       & Spleen         & Stomach        & P (M)    & GFLOPs        \\ \hline
				\multirow{4}{*}{CNNs} & R50 U-Net\cite{d9}       & 74.68          & 87.74          & 63.66          & \textbf{80.60} & \textbf{78.19} & \textbf{93.74} & 56.90          & 85.87          & 74.16          & -            & -            \\
				& R50 Att-UNet \cite{d9}   & 75.57          & 55.92          & 63.91          & 79.20          & 72.71          & 93.56          & 49.37          & 87.19          & 74.95          & -            & -            \\
				& U-Net\cite{d6}           & 76.85          & 89.07          & \textbf{69.72} & 77.77          & 68.60          & 93.43          & 53.98          & 86.67          & 75.58          & \textbf{40}  & 89           \\
				& Att-UNet\cite{d8}        & \textbf{77.77} & \textbf{89.55} & 68.88          & 77.98          & 71.11          & 93.57          & \textbf{58.04} & \textbf{87.30} & \textbf{75.75} & 45           & \textbf{84}  \\ \hline
				\multirow{4}{*}{\makecell[c]{3D\\Transformer}} & CoTr$^\star$\cite{d14}            & 72.60          & 83.27          & 60.41          & 79.58          & 73.01          & 91.93          & 45.07          & 82.84          & 64.67          & 42           & 377          \\
				& UNETR$^\star$\cite{d33}           & 79.56          & 89.99          & 60.56          & 85.66          & 84.80          & 94.46          & 59.25          & 87.81          & 73.99          & 92           & 86           \\
				& nnFormer$^\star$\cite{d34}        & 86.57          & 92.04          & 70.17          & 86.57          & 86.25          & 96.84          & \textbf{83.35} & 90.51          & \textbf{86.83} & 159          & 158          \\
				& D-Former$^\star$\cite{d23}        & \textbf{88.83} & \textbf{92.12} & \textbf{80.09} & \textbf{92.60} & \textbf{91.91} & \textbf{96.99} & 76.67          & \textbf{93.78} & 86.44          & \textbf{44}           & \textbf{54}           \\ \hline
				\multirow{7}{*}{\makecell[c]{2D\\Transformer}} & TransUnet$^\diamond$\cite{d9}       & 77.48          & 87.23          & 63.13          & 81.87          & 77.02          & 94.08          & 55.86          & 85.08          & 75.62          & 105          & 25           \\
				& Swin-UNet$^\diamond$\cite{d35}      & 79.13          & 85.47          & 66.53          & 83.28          & 79.61          & 94.29          & 56.58          & 90.66          & 76.60          & 41           & 12           \\
				& TransClaw U-Net\cite{d39} & 78.09          & 85.87          & 61.38          & 84.83          & 79.36          & 94.28          & 57.65          & 87.74          & 73.55          & -            & -            \\
				& LeVit-Unet-384$^\diamond$\cite{d36} & 78.53          & 87.33          & 62.23          & 84.61          & 80.25          & 93.11          & 59.07          & 88.86          & 72.76          & 52           & 26           \\
				& MT-UNet\cite{d37}         & 78.59          & 87.92          & 64.99          & 81.47          & 77.29          & 93.06          & 59.46          & 87.75          & 76.81          & -            & -            \\
				& MISSFormer\cite{d22}   & 81.96          & 86.99          & \textbf{68.65}          & 85.21          & 82.00         & 94.41          & 65.67          & \textbf{91.92}          & 80.81          & -            & -            \\
				& CA-GANformer$^\diamond$\cite{d38}    & 82.55          & 89.05          & 67.48 & 86.05          & 82.17          & \textbf{95.61} & 67.49          & 91.00          & \textbf{81.55} & -            & -            \\
				& \textbf{APFormer}   & \textbf{83.53} & \textbf{90.84} & 64.36          & \textbf{90.54} & \textbf{85.99} & 94.93          & \textbf{72.16} & 91.88 & 77.55          & \textbf{2.6} & \textbf{3.9} \\ \hline
			\end{tabular}
			\begin{tablenotes}
				\centering
				\item $\diamond$ represents the model needs pre-train on the large-scale dataset and $\star$ represents the model needs to train more than 1000 rounds.
			\end{tablenotes}
		\end{table*}
	}

	\section{Evaluation}
	
	\subsection{Datasets}	
	\subsubsection{ISIC 2018}
	
	This dataset is for skin lesion segmentation\cite{d41, d42}, consisting of 2596 images with pixel-level annotations. Following the setting in \cite{d29}, these images are split into 1815, 261, and 520 images for training, validation, and testing respectively.
	
	\subsubsection{Synapse}
	
	This dataset involves 30 cases of abdominal CT scans and each CT slice is annotated with 13 organs. Following the setting in \cite{d9} and the following transformer-based works \cite{d22,d35,d36,d37,d38,d39} on the Synapse dataset, 8 out of 13 organs are used for evaluation, and 18 cases are selected for training while the rest are built as the test set.  
	
	\subsection{Implementation Details}
	
	All learning frameworks are implemented based on PyTorch and trained for 400 rounds by an Adam optimizer with an initial learning rate of $10^{-4}$ and a batch size of 4. The parameter $g$ is initialized as $-2$ for stably pruning and is updated after 100 rounds. For data augmentation, methods including contrast adjustment, gamma augmentation, random rotation, and scaling are adopted.

	\subsection{Evaluation on ISIC 2018}
	
	\subsubsection{Learning Frameworks for Comparison}
	
	Four CNN-based methods including Att-UNet \cite{d8},  CPFNet\cite{d30}, DAGAN\cite{d31}, and CKDNet\cite{d32}, and three transformer-based approaches including SETR-PUP \cite{d5}, TransUnet \cite{d9}, and FAT-Net \cite{d29}, have been included for comparison. The experimental results of SETR-PUP and TransUnet are re-implemented according to the released source codes, while the rest quantitative results are reported by \cite{d29}.

	\subsubsection{Quantitative Results}
	
	According to the quantitative comparison results provided in Table~\ref{tab1}, the well-developed transformer-based approaches achieve better performance than those CNN-based approaches in skin lesion segmentation. Among these state-of-the-art approaches, FAT-Net achieves the best performance with a Dice score of 89.03$\%$. Compared to FAT-Net, despite a slight decrease in SE, APFormer achieves consistent performance improvements on Dice, IoU, ACC, and SP by an average increase of $1.67\%$, $2.32\%$, $0.72\%$, and $0.82\%$ respectively. More importantly, APFormer is much lighter than other approaches, which is more applicable in clinical scenarios.
	
	\subsubsection{Qualitative Results}
	
	Qualitative skin lesion segmentation results of various approaches including U-Net\cite{d6}, ResUNet\cite{d7}, Att-UNet\cite{d8}, SETR-PUP\cite{d5}, TransUnet\cite{d9}, FAT-Net\cite{d29}, and APFormer are presented in Fig.~\ref{fig5}. As shown in the first two rows, both CNN-based and hybrid-/transformer-based approaches suffer from local contextual similarity due to either limited receptive fields or superabundant dependencies. Comparatively, APFormer achieves the best segmentation performance based on the truly-valuable long-range dependency after pruning.
	
	\subsection{Evaluation on Synapse}
	
	\begin{figure*}[!t]
		\centerline{\includegraphics[width=1\textwidth]{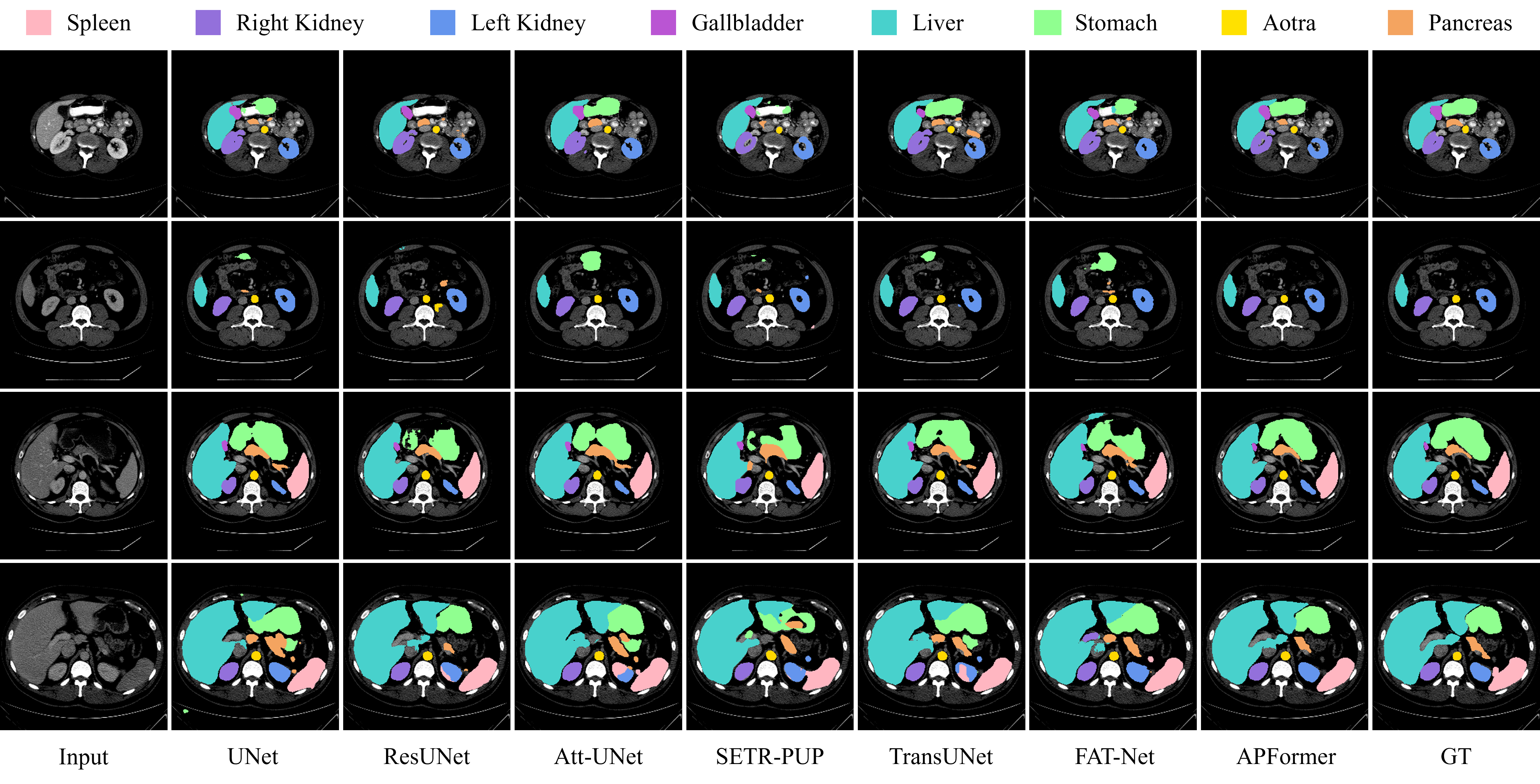}}
		\caption{Qualitative comparison results on the Synapse dataset. From left to right: the raw images, the segmentation results produced by U-Net \cite{d6}, ResUNet \cite{d7}, Att-UNet \cite{d8}, SETR-PUP \cite{d5}, TransUnet \cite{d9}, FAT-Net \cite{d29} and the proposed APFormer respectively, and the ground truth.}
		\label{fig6}
	\end{figure*}
	{\renewcommand{\arraystretch}{1.2}
		\begin{table*}[!t]
			\centering
			\caption{Comparison results of transformer pruning applied to different methods on the two datasets. $\alpha$, $\lambda$, and SA-GFLOPs are the query pruning rate, the dependency pruning rate, and the GFLOPs of self-attention modules respectively.}\label{tab4}
			\begin{tabular}{c|cccc|cccc}
				\hline
				\multirow{2}{*}{Method} & \multicolumn{4}{c|}{Synapse Dataset} & \multicolumn{4}{c}{ISIC 2018 Dataset} \\ \cline{2-9} 
				& Dice(\%)    & $\alpha$(\%) &$\lambda$(\%) & SA-GFLOPs & Dice(\%)    &$\alpha$(\%)  & $\lambda$(\%) & SA-GFLOPs \\ \hline
				SETR             & 73.22  & -       & -       & 12.89  & 88.03  & -        & -       & 12.89  \\
				AP-SETR               & 74.27  & -       & 84.36  & 12.89  & 88.97  & -        & 78.29   & 12.89  \\ \hline
				TransUNet         & 78.47  & -       & -       & 7.26  & 88.88  & -        & -       & 7.26  \\
				AP-TransUNet           & 79.82  & 91.17  & 97.31  & 5.44  & 89.60  & 69.27    & 97.44   & 7.17   \\ \hline
				APFormer-              & 82.71  & -       & -       & 1.63   & 89.94  & -        & -       & 1.63   \\
				APFormer               & 83.53  & 93.52  & 50.31  & 1.23   & 90.70  & 63.94   & 48.30  & 1.47    \\ \hline
			\end{tabular}
		\end{table*}
	}
	
	\subsubsection{Learning Frameworks for Comparison}
	
	CNN-based approaches including R50 U-Net\cite{d9}, R50 Att-UNet\cite{d9}, U-Net\cite{d6} and Att-UNet \cite{d8}, 2D transformer-based methods including TransUnet\cite{d9}, Swin-UNet\cite{d35}, TransClaw U-Net\cite{d39}, LeVit-Unet-384\cite{d36}, MT-UNet\cite{d37}, MISSFormer\cite{d22}, and CA-GANformer\cite{d38}, and 3D transformer-based methods including CoTr\cite{d14}, UNETR \cite{d33}, nnFormer \cite{d34}, and D-Former \cite{d23} are used for comparison.
	
	\subsubsection{Quantitative Results}
	
	Quantitative comparison results of different methods on the Synapse dataset are summarized in Table~\ref{tab2}. Compared to the CNN-based methods, the transformer-based methods generally achieve better segmentation results due to the modeling of long-range dependency, among which nnFormer and D-Former achieve the best overall performance at the cost of higher computational complexity in GFLOPs. Comparatively, 2D transformer-based methods can achieve satisfactory performance with much lower complexity. CA-GANformer has the best performance with an average Dice of $82.55\%$ and Swin-UNet owns the lowest computational complexity. Compared to the state-of-the-art 2D transformer-based methods, APFormer achieves the best overall performance with the lowest computational complexity. Specifically, APFormer outperforms Swin-UNet with an average increase of 4.40\% in Dice and 3x lower GFLOPs. Compared to D-Former, APFormer is about 17x lighter with 14x lower computational complexity in GFLOPs, which significantly improves the feasibility for clinical deployment.
	
	\subsubsection{Qualitative Results}
	
	Qualitative segmentation results of different methods on the Synapse dataset are illustrated in Fig.~\ref{fig6}. As analyzed in \cite{d9, d10}, the CNN-based methods tend to make segmentation errors in local similarity regions due to their inner inductive biases. Theoretically, the hybrid-/transformer-based methods can overcome this issue by modeling long-range dependency. However, perception redundancy as analyzed in Section II would negatively affect the segmentation performance. These reasons explain why transformer pruning significantly improves the segmentation results.
	
	\section{Discussion}
	
	Ablation studies are conducted to validate the generalization ability and effectiveness of APFormer in the following.
	
	\subsection{Generalization of Transformer Pruning}
	
	\begin{figure*}[!t]
		\centerline{\includegraphics[width=1\textwidth]{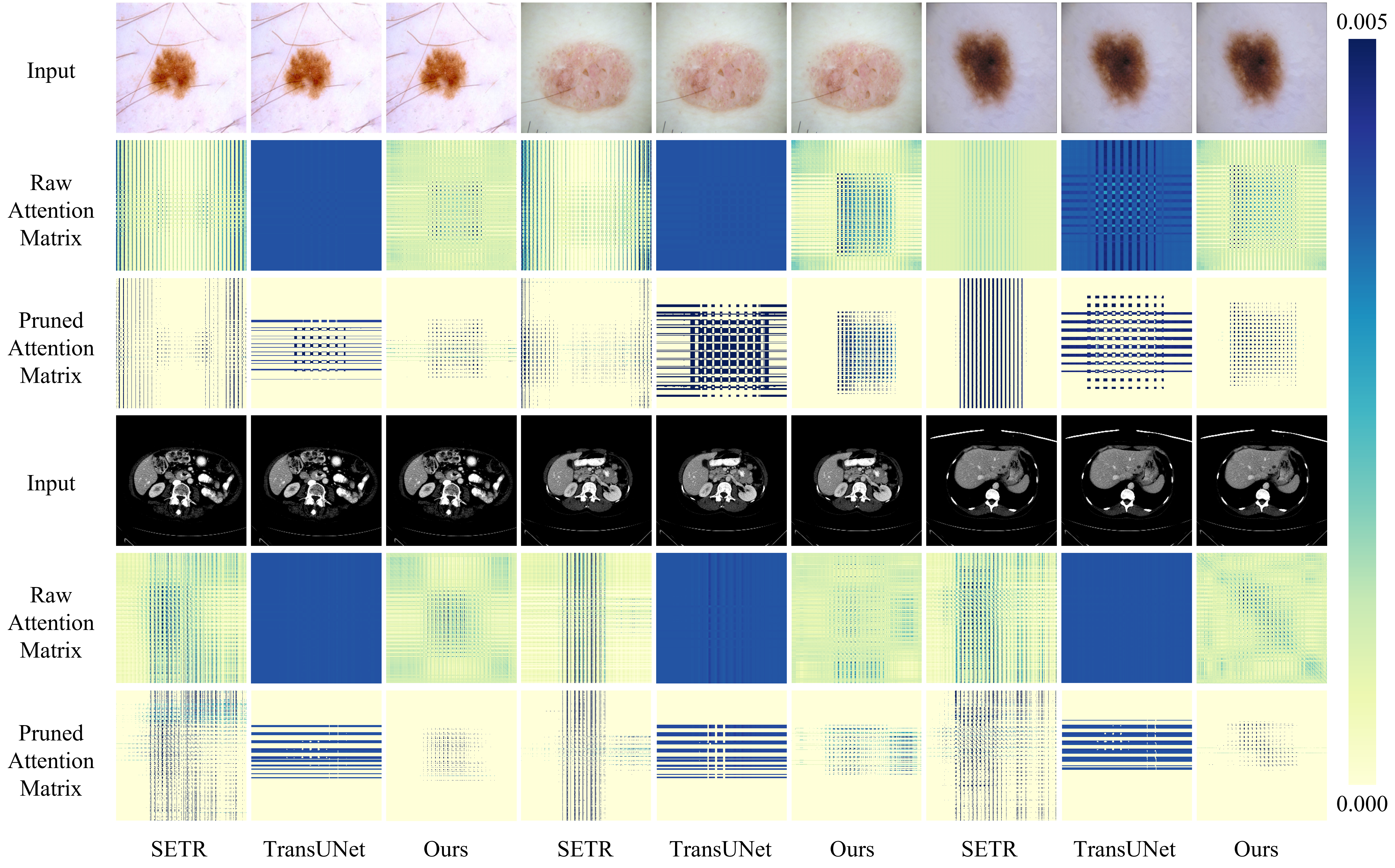}}
		\caption{Visual results of self-attention matrix changes by applying transformer pruning in SETR, TransUnet, and APFormer on the Synapse and ISIC 2018 datasets.}
		\label{fig8}
	\end{figure*}
	Transformer pruning is further coupled with other hybrid-/transformer-based approaches for evaluation. Based on the availability of source code, SETR and TransUnet were re-implemented for comparison. For SETR, only dependency-wise pruning is applied as it directly feeds raw images to the transformers, making it impossible to generate the learnable gate scores for query-wise pruning. For TransUnet, we apply both query-wise and dependency-wise pruning. 
	
	Quantitative comparison results of different methods before and after pruning are summarized in Table~\ref{tab4}. With dependency-wise pruning, the segmentation performance of SETR can be improved by $1.05\%$ and $0.94\%$ respectively in Dice on the Synapse and the ISIC 2018 datasets. By introducing both query-wise and dependency-wise pruning, on the Synapse and the ISIC 2018 datasets, the computational complexity of TransUNet can be reduced by $25.07\%$ and $1.24\%$ with an average increase of $1.35\%$ and $0.72\%$ in Dice respectively. One interesting observation is that SETR and TransUNet benefit more from pruning compared to APFormer-, validating the value and architecture independence of the proposed adaptive transformer pruning.
	
	{\renewcommand{\arraystretch}{1.2}
		\begin{table}[t]
			\centering
			\caption{Ablation study on different component combinations of APFormer on the ISIC 2018 dataset. APFormer- represents that APFormer without adaptive pruning.}\label{tab3}
			\begin{tabular}{cccccc}
				\hline
				Method            & Dice (\%)       & IoU (\%)        & ACC (\%)        & SE (\%)         & SP (\%)         \\ \hline
				baseline & 88.89          & 81.90          & 96.15          & 89.52          & 97.63          \\
				+ SSA              & 89.39          & 82.45          & 96.12          & 90.31          & 97.90          \\
				+ GRPE              & 89.63          & 82.75          & 96.10          & 89.91   & \textbf{97.94}          \\
				+ Pruning              & 89.29          & 82.28          & 96.09          & 90.93 & 97.42          \\
				APFormer-              & 89.94          & 83.15          & 96.24          & \textbf{91.66} & 96.87          \\
				\textbf{APFormer}  & \textbf{90.70} & \textbf{84.34} & \textbf{96.50} & 90.93          & 97.81 \\ \hline
			\end{tabular}
		\end{table}
	}
	To analyze the effect of the proposed transformer pruning method, we visualize the changes in self-attention matrices before and after pruning. As shown in Fig. \ref{fig8}, in the raw attention matrices, though the dependency of each query would focus on some useful patches, redundant dependency from irrelevant patches generates the long tail effect and introduces non-negligible noise. After pruning, dependency relationships become ``cleaner'' and more distinguishable, leading to performance improvement. Moreover, for those easy-to-identify patches, building global dependency is worthless, incurring unnecessary computations. Therefore, pruning redundant patches can effectively reduce computational complexity. 
	
	\subsection{Effectiveness of Each Component in APFormer}
	
	The three components in APFormer, namely SSA, GRPE, and adaptive pruning, are separately introduced to the backbone network and trained on the ISIC 2018 dataset for evaluation. As presented in Table~\ref{tab3}, each component included contributes to performance improvement, indicating the effectiveness of each individual component. Furthermore, compared with directly applying adaptive pruning to the baseline, introducing SSA and GRPE can not only improve the segmentation performance but also increase the pruning effect from 0.4$\%$ to 0.76$\%$ in Dice improvements, which is consistent with the analysis in Section II. 
	
	\begin{figure}[!t]
		\centerline{\includegraphics[width=1\columnwidth]{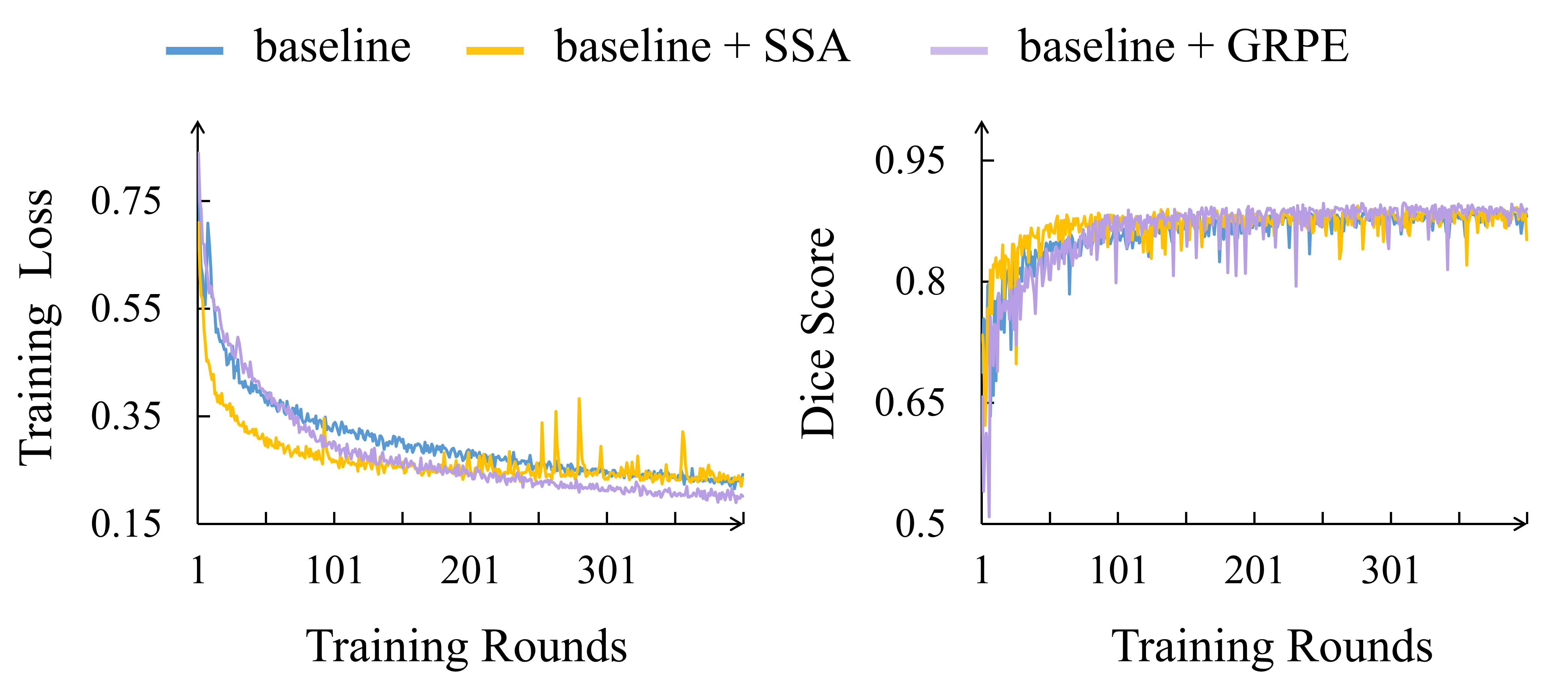}}
		\caption{Training loss curves and Dice curves of the baseline model coupled with SSA and GRPE separately.}
		\label{fig7}
	\end{figure}
	Training curves of the baseline network with SSA and GRPE separately are provided in Fig. \ref{fig7}. The main contribution of SSA is to speed up the convergence, which is quite useful for training on small-scale datasets. Comparatively, GRPE provides more useful position information, leading to better convergence performance and segmentation results.	
	
	\section{Conclusion}
	
	In this paper, we propose a lightweight yet effective hybrid CNN-Transformer framework APFormer for medical image segmentation from the perspective of transformer pruning. Specifically, self-supervised attention is proposed to speed up the convergence of transformers which is helpful to alleviate the data dependence problem. Then, Gaussian-prior relative position embedding is designed to effectively supply the position information to transformers. After that, we perform query-wise and dependency-wise pruning sequentially for not only redundancy reduction but also performance improvement.	
	Experimental results on two public datasets demonstrate the superior performance of APFormer compared to the state-of-the-art 2D/3D methods with significantly lower computational complexity. Furthermore, we demonstrate, through extensive evaluation, that adaptive pruning is architecture-independent and applicable to other hybrid-/transformer-based frameworks for performance improvement. 
	We believe the idea of transformer pruning would inspire future work on developing ultra-light and even better-performing transformers for medical image segmentation.

\end{document}